\documentclass{article}
\usepackage{spconf,amsmath,graphicx}
\usepackage{array}
\usepackage{tabularx}
\usepackage{booktabs}
\usepackage{multirow}
\usepackage{lipsum}

\newcommand{\mytilde}{\raise.2ex\hbox{$\scriptstyle\sim$}}
\newcommand{\ie}{\textit{i}.\textit{e}.\ }

\makeatletter
\DeclareRobustCommand{\cev}[1]{%
  \mathpalette\do@cev{#1}%
}
\newcommand{\do@cev}[2]{%
  \fix@cev{#1}{+}%
  \reflectbox{$\m@th#1\vec{\reflectbox{$\fix@cev{#1}{-}\m@th#1#2\fix@cev{#1}{+}$}}$}%
  \fix@cev{#1}{-}%
}
\newcommand{\fix@cev}[2]{%
  \ifx#1\displaystyle
    \mkern#23mu
  \else
    \ifx#1\textstyle
      \mkern#23mu
    \else
      \ifx#1\scriptstyle
        \mkern#22mu
      \else
        \mkern#22mu
      \fi
    \fi
  \fi
}

\makeatother

\title{Contextualized Multimodal Representations}
\title{Multimodal Embeddings from Language Models}
\name{
Shao-Yen Tseng, 
Panayiotis Georgiou, 
Shrikanth Narayanan
}
\address{
Department of Electrical and Computer Engineering \\
University of Southern California \\
Los Angeles, CA, USA \\
}
\begin{document}
%
\maketitle
\begin{abstract}

Word embeddings such as ELMo have recently been shown to model word semantics with greater efficacy through contextualized learning on large-scale language corpora, resulting in significant improvement in state of the art across many natural language tasks.
In this work we integrate acoustic information into contextualized lexical embeddings through the addition of multimodal inputs to a pretrained bidirectional language model.
The language model is trained on spoken language that includes text and audio modalities. 
The resulting representations from this model are multimodal and contain paralinguistic information which can modify word meanings and provide affective information.
We show that these multimodal embeddings can be used to improve over previous state of the art multimodal models in emotion recognition on the CMU-MOSEI dataset.

\end{abstract}
%
%
\section{Introduction}
\label{sec:intro}

Acoustic and visual elements in human communication, such as intonation or facial expressions, infuses semantic content with additional paralinguistic cues which may modify intent and can convey affective meaning more clearly if not exclusively \cite{rodero2011intonation}.
For this reason many work have proposed multimodal systems which integrate information from multiple modalities to improve natural language understanding.
This effort encompasses research in many applications such as human robot interfaces \cite{meszaros2017compensating, maurtua2017natural}, video summarization \cite{nihei2017predicting, nihei2018fusing, palaskar2019multimodal}, dialogue systems \cite{liao2018knowledge, hori2019end}, and emotion and sentiment analysis \cite{hu2018multimodal, yoon2018multimodal, liang2018multimodal}. 

The study of multimodal fusion in affective systems is a prevalent and important topic. 
This follows from the fact that human behavioral expression is fundamentally a multifaceted phenomena that spans over multiple modalities \cite{narayanan2013behavioral, d2015review} and can be more accurately identified through multimodal classifiers \cite{poria2017review}. 
Another factor is the importance of affective information as an intermediate step in a variety of downstream tasks.
Examples of which include the use of these human behavioral states in language modeling \cite{shivakumar2019behavior}, dialogue systems \cite{pittermann2010emotion, bertero2016real}, and video summarization \cite{singhal2018summarization}. 

Many multimodal systems for recognition of sentiment, emotion, and behaviors have been proposed in prior work. 
In feature-level fusion, Tzirakis et al. \cite{tzirakis2017end} combined auditory and visual modalities by extracting features from convolutional neural networks (CNN) on each modality which were then concatenated as input to an LSTM network. 
Hazarika et al. \cite{hazarika2018self} proposed the use of a self-attention mechanism to assign scores for weighted combination of modalities.
Other work has applied multimodal integration using late fusion methods \cite{tseng2018honey, blanchard2018getting}.

For deeper integration between modalities many work have proposed the use of multimodal neural architectures. 
Lee et al. \cite{lee2018convolutional} proposed the use of an attention matrix calculated from speech and text features to selectively focus on specific regions of the audio feature space.
The memory fusion network was introduced by Zadeh et al. \cite{zadeh2018memory} which accounted for intra- and inter-modal dependencies across time. 
Akhtar et al. \cite{akhtar2019multi} proposed a contextual inter-modal attention network which leveraged sentiment and emotion labels in a multi-task learning framework.

The strength of deep models arises from the ability to internally learn meaningful representations of features from multiple modalities.
This is learned implicitly by the model through the course of training on respective datasets \cite{bengio2013representation}. 
In this work we propose a model to explicitly learn informative joint representations of speech and text.
This is achieved by modeling the dynamics between lexical content and paralinguistics from audio through a language modeling task on spoken language. 
We augment a bidirectional language model (biLM) with word-aligned acoustic features and optimize the model using large-scale text corpora followed by spoken articles. 
The internal states of this biLM aren't representative of a specific task but rather models the intricacies of human communication through speech and language. 
We show the effectiveness of representations extracted from this model in capturing multimodal information by evaluating in the task of emotion recognition.
Through the use of these representation we improve the state of the art in emotion recognition on the CMU-MOSEI dataset.

\section{Related Work}

Lexical representations such as ELMo \cite{peters2018deep} and BERT \cite{devlin2019bert} have recently been shown to model word semantics and syntax with greater efficacy.
This is achieved through contextualized learning on large-scale language corpora which allows internal states of the model to capture both the complex characteristics of word use as well as polysemy due to different contexts.
The integration of these word embeddings into downstream models have improved the state of the art in many NLP tasks through their rich representation of language use.

To learn representations from multimodal data Hsu et al. \cite{hsu2018disentangling} proposed the use of variational autoencoders to encode inter- and intra-modal factors into separate latent variables.
Later, Tsai et al. \cite{tsai2018learning} factorized representations into multimodal discriminative and modality-specific generative factors using inference and generative networks. 
During the course of writing this paper Rahman et al. \cite{rahman2019m} concurrently proposed the infusion of multimodal information into the BERT model. 
There the authors combined the generative capabilities of the BERT model with a sentiment prediction task to allow the model to implicitly learn rich multimodal representations through a joint generative-discriminative objective. 

In this work we propose to explicitly learn multimodal representations of spoken words by augmenting the biLM model in ELMo with acoustic information. 
This is motivated from how humans integrate acoustic characteristics in speech to interpret the meaning and intent of lexical content from a speaker. 
Our work differs from prior work in that we do not include or target any discriminative objectives and instead rely on the generative task of language modeling to learn meaningful multimodal representations.
We show how this model can be easily trained with large-scale unlabeled data and also demonstrate how potent multimodal embeddings from this model are in tasks such as emotion recognition. 

\section{Multimodal Embeddings}

\begin{figure*}[t]
    \centering
    \includegraphics[width=0.93\linewidth]{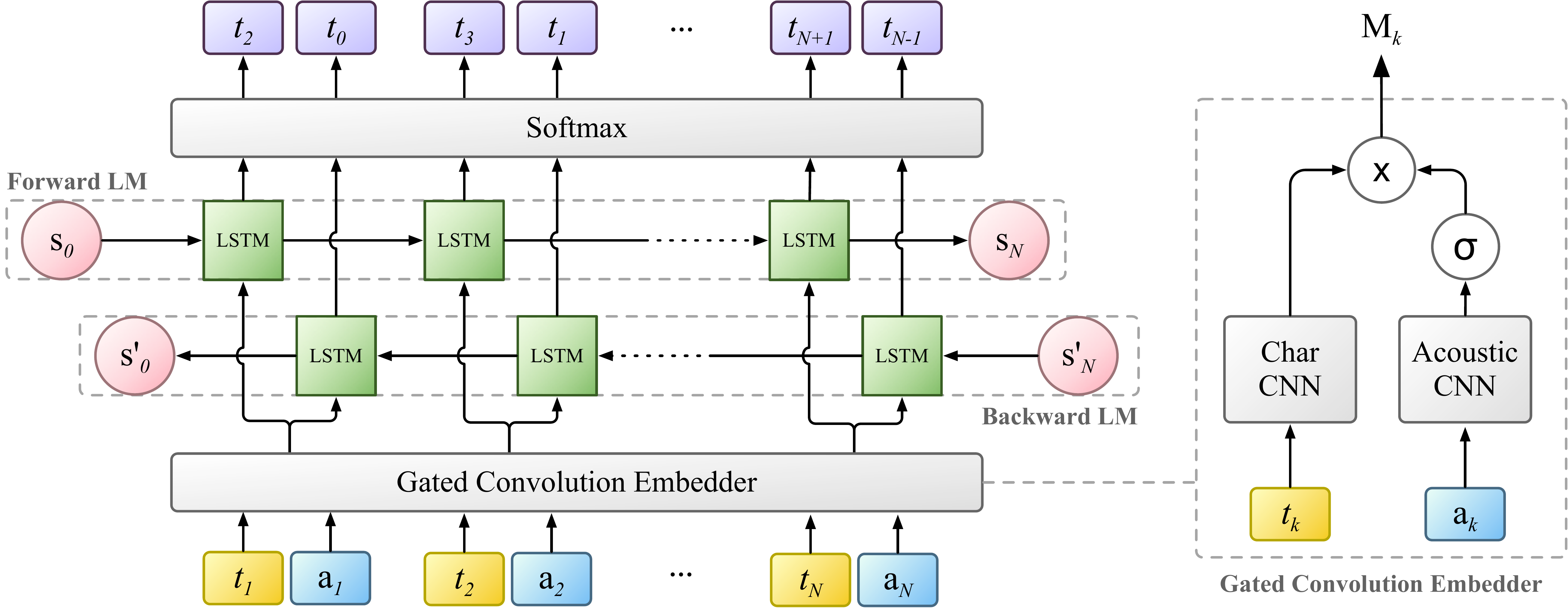}
    \caption{Architecture of the multimodal bidirectional language model.}
    \label{fig:melmo}
\end{figure*}

In this section we describe the network architecture of the bidirectional language model with acoustic information that generates the multimodal embeddings.
The biLM comprises two layers of bidirectional LSTMs which operate over lexical and audio embeddings.
The lexical and audio embeddings are calculated from respective  convolutional layers and combined using a sigmoid-gating function.
Multimodal embeddings are then computed using a linear function over the internal states of the recurrent layers. 
The overall architecture of the mutlimodal biLM is shown in Figure \ref{fig:melmo}.

\subsection{Bidirectional language model}

A language model (LM) computes the probability distribution of a sequence of words by approximating it as the product of conditional probabilities of each word given previous words.  
This has been implemented using neural networks in many prior work yielding state of the art results \cite{merity2018regularizing, gong2018frage}. 
In this work we applied the biLM model used in ELMo, which is similar to the character-level RNN-LM described by J\'ozefowicz et al. \cite{jozefowicz2016exploring} and Kim et al. \cite{kim2016character}. 

The biLM is composed of a forward and backward LM each implemented by a two-layer LSTM. 
The forward LM predicts the probability distribution of the next token given past context while the backward LM predicts the probability distribution of the previous token given future context.
Each LM operates on the same input, which is a token embedding of the current token calculated through a character-level convolutional neural network (CharCNN).
A softmax layer is used to estimate token probabilities from the output of the   two-layer LSTM in the LMs. 
The parameters of the softmax layer are shared between the LMs in both directions. 

Different from ELMo our input to the biLM includes acoustic features in additional to word tokens. 
Now the forward LM aims to model, at each timestep, the conditional probability of the next token $ t_{k+1} $ given the current token $ t_{k} $, acoustic features $ \mathbf{a}_{k} $, and previous internal states of the two-layer LSTM $ \vec{\mathbf{s}}_{k-1} $:
$$ P( t_{k+1} \mid t_k, \mathbf{a}_k, \vec{\mathbf{s}}_{k-1} ) $$
The backward LM operates similarly but predicts the previous token $t_{k-1} $ given the current token $ t_{k} $, acoustic features $ \mathbf{a}_{k} $, and internal states resulting from future context $ \cev{\mathbf{s}}_{k+1} $.

\subsection{Acoustic convolution layers}
In our implementation time-aligned acoustic features of each word are provided in adjunct to word tokens.  
We build on ELMo and add additional convolutional layers to calculate acoustic embeddings from the acoustic features. 
The convolutional layers provide a feature transformation of the acoustic features which we combine with token embeddings using a gating function.

Due to the varying lengths of words in time, the acoustic features are first padded to a fixed size in the temporal dimension before being passed to the CNN. 
Each convolution layer in the CNN comprises a 1-D convolution layer followed by a max-pooling layer.  
Finally, the feature map is projected to the same dimension size as token embeddings to allow for element-wise combination.

\subsection{Multimodal ELMo}
\label{sec:melmo}

We combine token and acoustic embeddings using a sigmoid gating function:
$$ \mathbf{M}_k = \mathbf{U}(t_k) \odot \sigma(\mathbf{V}(\mathbf{a}_k))$$
where $\mathbf{U}$ and $\mathbf{V}$ are the embeddings calculated from the token and corresponding acoustic features, respectively, $\sigma$ is the sigmoid function, and $\odot$ represents element-wise multiplication.  
The resulting multimodal embeddings are used as input to the forward and backward LM. 

The sigmoid gate scales the token embedding based on corresponding acoustic features of the word. 
This serves as a modifier of semantic meaning using paralinguistic information which we hypothesize will be useful in capturing affective expressions in downstream tasks.

Word embeddings are extracted for use in downstream models in a similar fashion to ELMo. That is, we compute a task-specific weighted sum of all LSTM outputs as each word embedding.
We adopt the use of sentence embeddings in downstream models and additionally average all the word embeddings in a sentence. 
The final multimodal ELMo (M-ELMo) sentence embedding is given as
$$ \mathbf{M{\text -}ELMo} = \gamma \frac{1}{N} \sum_{j}^{L}c_j \sum_{k}^{N} \mathbf{h}_{k,j}$$
where $\mathbf{h}_{k,j}$ are the concatenated outputs of LSTMs in both directions at the $j^{\mathrm{th}}$ layer for the $k^{\mathrm{th}}$ token.
Values $\{c_j\}$ are softmax-normalized weights and $\gamma$ is a scalar value, all of which are tunable parameters in the downstream model. 

\section{Experimental Setup}

\begin{table*}[t]
    \centering
    \begin{tabular}{lcccccccccccccc}
        \toprule
        &  \multicolumn{2}{c}{Anger} & \multicolumn{2}{c}{Disgust} & \multicolumn{2}{c}{Fear} 
            & \multicolumn{2}{c}{Happy} & \multicolumn{2}{c}{Sad}  & \multicolumn{2}{c}{Surprise} & \multicolumn{2}{c}{\textbf{Average$^\ddagger $}}\\
        \textbf{Method} & WA & F1 & WA & F1 & WA & F1 & WA & F1 & WA & F1 & WA & F1 & \textbf{WA} & \textbf{F1}  \\ 
        
        \toprule
        
        \textbf{(A + L + V)} \\
        Graph-MFN \cite{zadeh2018multimodal} &
        62.6 & 72.8 & 69.1 & 76.6 & 62.0 & \textbf{89.9} & 66.3 & \textbf{66.3} & 60.4 & 66.9 & 53.7 & 85.5 & 62.3 & 76.3 \\
        
        CIM-Att-STL \cite{akhtar2019multi} &
        64.5  & \textbf{75.6} & 72.2 &  81.0 & 51.5 & 87.7 & 61.6 & 59.3 & \textbf{65.4} &  67.3 & 53.0 & \textbf{86.5} & 61.3 & 76.2 \\
        
        \midrule
        \textbf{(A + L)} \\
        CIM-Att-STL \cite{akhtar2019multi} &
        -  & - & - & - & -  & - & - & - & -  & - & - & - & 59.6 & 76.8 \\
        
        
        M-ELMo + NN $^\S $ &
        \textbf{65.8} & 74.7 & \textbf{74.2} & \textbf{81.7} & \textbf{63.2} & 85.1 & \textbf{67.0} & 65.2 & 63.1 & \textbf{72.0} & \textbf{63.8} & 83.3 & \textbf{66.2} & \textbf{77.0} \\
        
        
        \bottomrule
    \end{tabular}
    \caption{Emotion recognition results on CMU-MOSEI test set for various multimodal models.
    Modalities: acoustic (A), lexical (L), visual (V). 
    Values are taken from their respective sources.  
    $^\S $Average across ten runs. 
    $^\ddagger $Average across six emotions.}
    \label{tab:res_mosei}
\end{table*}

\subsection{Pre-training the multimodal biLM}

The multimodal biLM is pre-trained in two stages. 
In the first stage the lexical components of the biLM are optimized prior to the inclusion of acoustic features. 
This is achieved by training with a text corpus and fixing the acoustic input as zero. 
We use the 1 Billion Word Language Model Benchmark \cite{chelba2014one} for this purpose and train the biLM for 10 epochs. 
After training, the model achieves perplexities of around 35 which is similar to values reported in \cite{peters2018deep}.

In the second stage of training we optimize the biLM using the multimodal dataset CMU-MOSEI \cite{zadeh2018multimodal} (described in Section \ref{sec:setup_eval}).
In our experiments we use only the text and audio of segments in the training split of the dataset to train the model. 
In terms of word count CMU-MOSEI is much smaller than the 1-billion word LM benchmark, therefore to prevent overfitting we reduce the learning rate used in the previous stage by a factor of 10 and train for an additional 5 epochs. 

\subsection{Features}

Since a CharCNN is used as the lexical embedder, input words to the biLM are first transformed into a character map and padded to a fixed length. 
The character-level representation of each word is then given as a $c \times l_c$ matrix, where $c$ is the dimension size of the character embedding and $l_c$ is the maximum number of characters in a word.

Acoustic features were extracted from each recording at 10ms frame intervals using the COVAREP software version 1.4.2 \cite{degottex2014covarep}.
There are 74 features in total and include, among others, pitch, voiced/unvoiced segment features, mel-frequency cepstral coefficients, glottal flow parameters, peak slope parameters, and harmonic model parameters.

The acoustic features are aligned with word timings to provide acoustic information for each word.  
Since the time duration varies between words we pad the number of acoustic frames per token to a fixed length.
Thus, word-aligned acoustic features are given as a $d \times l_a$ matrix, where $d$ is the number of acoustic features and $l_a$ is the maximum number of acoustic frames in a word.

        
        

\subsection{Emotion recognition as a downstream task}
\label{sec:setup_eval}

After pre-training, the multimodal biLM is used to extract multimodal sentence embeddings for use in downstream models.
In our experiments we adopt emotion recognition as the downstream task and evaluate on the CMU-MOSEI dataset.

\textbf{CMU-MOSEI} contains 23,453 single-speaker video segments from YouTube which have been manually transcribed and annotated for sentiment and emotion.
Emotions are annotated on a [0,3] Likert scale and include those such as  \textit{happiness}, \textit{sadness}, \textit{anger}, \textit{fear}, \textit{disgust}, and \textit{surprise}. 
We binarize these annotations to arrive at class labels by predicting the presence of emotions, \ie any emotion with a rating greater than one. 
Since video segments have ratings for all emotions this becomes a multi-label classification task. 


As our goal is to evaluate the efficacy of the multimodal sentence embeddings we used a simple feedforward neural network for emotion recognition.
The network inputs the multimodal sentence embeddings and predicts the presence of each emotion. 
The tunable parameters described in Section \ref{sec:melmo} are also included in this network. 
We trained the network using data from the training split provided in the dataset and validated using the validation split. 
We also used the validation split as a development set in choosing hyper-parameters of the network. 


\subsection{Evaluation methods}

We evaluated the emotion recognition model using weighted accuracy (WA) and F1 score on each emotion. 
Weighted accuracy, as used in \cite{zadeh2018multimodal}, is equivalent to the macro-average recall value. 
We also averaged the metrics across all emotions to obtain an average WA and F1 score.

The downstream model was trained for 30 epochs and separately optimized on WA and F1 score using the validation set. 
We randomly initialized each downstream model ten times and a best model was selected based on the average scores on validation over the ten runs.
The final model was a neural network with two hidden layers using ReLU activation functions.

Due to the lack of work that only focuses on text and audio, we compared with models that also considers the visual modality.  
We compared our performance with two recent state of the art emotion recognition models on CMU-MOSEI.
Specifically, these are the graph Memory Fusion Network (Graph-MFN) \cite{zadeh2018multimodal} and the contextual inter-modal attention framework (CIM-Att) \cite{akhtar2019multi}.
To match learning conditions, we compared with the single task learning (STL) model of \cite{akhtar2019multi} where only emotion labels are used in training.

\section{Results}

The results of the final model averaged across ten runs are shown in Table \ref{tab:res_mosei}.
Our simple feedforward neural network using multimodal embeddings achieved state of the art results in terms of average WA and F1 over all emotions at 66.2\% and 77.0\%, respectively.
On individual emotions our model yielded comparable to improved results over state of the art. 
Specifically, we observed improvements in the weighted accuracy of all emotions except \textit{sad} as well improvements in F1 score of \textit{disgust} and \textit{sad}.

\section{Conclusion}

In this work we proposed a method for extending ELMo word embeddings to include acoustic information. 
We used convolutional layers over word-aligned acoustic features to calculate acoustic embeddings which we then combined with token embeddings in ELMo using a sigmoid gating function. 
The model was trained on a language modeling task, first with a text corpus followed by inclusion of audio from a multimodal dataset. 
We then showed the effectiveness of sentence embeddings extracted from this multimodal biLM in emotion recognition. 
The results are surprising given that our downstream model using a neural network with two hidden layers outperformed state of the art architectures. 
This demonstrates how well the multimodal embeddings have captured inter- and intra-modal dynamics in spoken language.

\vfill
\clearpage
\pagebreak

\bibliographystyle{IEEEbib}
\bibliography{strings,refs}

\end{document}